\documentclass[10pt,twocolumn,letterpaper]{article}

\usepackage{cvpr}              
\usepackage{subcaption}
\definecolor{cvprblue}{rgb}{0.21,0.49,0.74}
\usepackage[pagebackref,breaklinks,colorlinks,allcolors=magenta]{hyperref}
\usepackage{makecell}
\def\paperID{*****} 
\def\confName{CVPR}
\def\confYear{2025}

\title{NTIRE 2026 Challenge on Short-form UGC Video Restoration in the Wild with Generative Models: Datasets, Methods and Results}

\author{Xin Li\textsuperscript{$\dagger$} \quad Jiachao Gong\textsuperscript{$\dagger$} \quad Xijun Wang\textsuperscript{$\dagger$} \quad Shiyao Xiong\textsuperscript{$\dagger$} \quad Bingchen Li \textsuperscript{$\dagger$} \\ Suhang Yao\textsuperscript{$\dagger$} \quad Chao Zhou\textsuperscript{$\dagger$} \quad Zhibo Chen\textsuperscript{$\dagger$} \quad Radu Timofte\textsuperscript{$\dagger$} \\
Yuxiang Chen \quad Shibo Yin \quad Yilian Zhong \quad Yushun Fang \quad Xilei Zhu \quad Yahui Wang\quad Chen Lu \\
Meisong Zheng\quad Xiaoxu Chen\quad Jing Yang\quad Zhaokun Hu \quad Jiahui Liu\\ Ying Chen\quad Haoran Bai\quad Sibin Deng\quad Shengxi Li\quad Mai Xu \\ 
 Junyang Chen\quad Hao Chen\quad Xinzhe Zhu\quad Fengkai Zhang\quad Long Sun\\ Yixing Yang\quad Xindong Zhang\quad Jiangxin Dong\quad Jinshan Pan \\
 Jiyuan Zhang \quad Shuai Liu\quad Yibin Huang\quad Xiaotao Wang\quad Lei Lei \\
 Zhirui Liu \quad Shinan Chen\quad Shang-Quan Sun\quad Wenqi Ren \\
 Jingyi Xu \quad Zihong Chen\quad Zhuoya Zou \\
 Xiuhao Qiu \quad  Jingyu Ma\quad  Huiyuan Fu\quad  Kun Liu\quad  Huadong Ma\\  Dehao Feng\quad  Zhijie Ma\quad  Boqi Zhang\quad  Jiawei Shi\quad  Hao Kang  \\
 Yixin Yang\quad Yeying Jin\quad Xu Cheng\\
  Yuxuan Jiang\quad  Chengxi Zeng\quad  Tianhao Peng\quad  Fan Zhang\quad  David Bull \quad
  Yanan Xing  \\
  Jiachen Tu \quad Guoyi Xu\quad Yaoxin Jiang\quad Jiajia Liu\quad Yaokun Shi \\
  Wei Zhou \quad Linfeng Li \quad Hang Song \quad Qi Xu \quad Kun Yuan \quad Yizhen Shao \quad Yulin Ren
}

\begin{document}
\maketitle
\renewcommand{\thefootnote}{}
\footnotetext{$^{*}$X. Li(\textcolor{magenta}{xin.li@ustc.edu.cn}), J. Gong, X. Wang, S. Xiong, B. Li, S. Yao, C. Zhou, Z. Chen, R. Timofte are the challenge organizers. K. Yuan, Y. Shao and Y. Ren are the supporting staff of this challenge. (Corresponding author: Xin Li)} 

\footnotetext{The other authors are participants of the NTIRE 2026 Challenge on Short-form UGC Video Restoration in the Wild with Generative Models.}
\footnotetext{The Competition website:~\url{https://www.codabench.org/competitions/13256/}}
\footnotetext{The NTIRE2026 website:~\url{https://cvlai.net/ntire/2026/}}

\begin{abstract}
This paper presents an overview of the NTIRE 2026 Challenge on Short-form UGC Video Restoration in the Wild with Generative Models. This challenge utilizes a new short-form UGC (S-UGC) video restoration benchmark, termed KwaiVIR, which is contributed by USTC and Kuaishou Technology. It contains both synthetically distorted videos and real-world short-form UGC videos in the wild. For this edition, the released data include 200 synthetic training videos, 48 wild training videos, 11 validation videos, and 20 testing videos. The primary goal of this challenge is to establish a strong and practical benchmark for restoring short-form UGC videos under complex real-world degradations, especially in the emerging paradigm of generative-model-based S-UGC video restoration. This challenge has two tracks: (i) the primary track is a subjective track, where the evaluation is based on a user study; (ii) the second track is an objective track. These two tracks enable a comprehensive assessment of restoration quality. In total, 95 teams have registered for this competition. And 12 teams submitted valid final solutions and fact sheets for the testing phase. The submitted methods achieved strong performance on the KwaiVIR benchmark, demonstrating encouraging progress in short-form UGC video restoration in the wild. The project is publicly available at~\url{https://github.com/lixinustc/KVQE-Challenge-CVPR-NTIRE2026}.
\end{abstract}
    
\section{Introduction}
\label{sec:intro}

Short-form user-generated content videos (\ie, S-UGC videos)~\cite{ntire2025shortugc,ntire2025shortugc_data,KSVQE,li2024ntire} have become one of the most important forms of visual media on modern mobile platforms, like Kwai and TikTok platforms. Compared with traditional professionally produced videos, S-UGC videos are typically captured by ordinary users under unconstrained conditions and then transmitted through complex in-platform processing pipelines. As a result, the visual quality of S-UGC videos is often degraded by a mixture of factors, including consumer-level acquisition limitations, compression and transcoding artifacts, unstable motion, and other real-world distortions. These degradations not only reduce perceptual quality, but also pose substantial challenges to practical restoration systems deployed in real applications.

Video restoration~\cite{chan2022investigating,zhang2024realviformer,li2026test,li2025hybrid} for S-UGC content is fundamentally different from conventional restoration settings. Existing benchmarks and methods are often developed under relatively controlled assumptions, where degradations can be synthetically modeled or restricted to specific categories. In contrast, real-world S-UGC videos usually exhibit diverse and entangled distortions, making restoration much more difficult in the wild. In addition to distortion removal, practical methods~\cite{zhuang2025flashvsr} are also expected to preserve temporal coherence, maintain perceptual realism, and generalize well across highly diverse content scenarios. These characteristics make S-UGC video restoration a challenging yet important research problem.

Recent advances in generative models have opened up new opportunities for video restoration~\cite{uav,wang2025liftvsr,dove,zhuang2025flashvsr,li2025diffusion}. By leveraging powerful generative priors~\cite{yang2024cogvideox,StableSR,ren2024moe}, modern restoration methods are able to recover visually plausible details and produce perceptually appealing results even under complex degradations that are difficult to characterize explicitly. Nevertheless, despite this rapid progress, there is still a lack of dedicated benchmarks for generative restoration~\cite{li2026test} of S-UGC videos in realistic platform scenarios. In particular, a practical benchmark should support both fidelity-oriented evaluation on paired synthetic data and perceptual evaluation on real-world videos, while also accounting for temporal consistency in restored outputs.

To promote research in this direction, we organize the NTIRE 2026 Challenge on Short-form UGC Video Restoration~\cite{zhuang2025flashvsr} in the Wild with Generative Models~\cite{yang2024cogvideox}. This challenge is built upon the newly introduced \textit{KwaiVIR} benchmark, which contains both synthetically distorted video pairs and real-world S-UGC videos collected in the wild. By combining these two data sources within a unified challenge framework, the benchmark is designed to better reflect the practical requirements of restoration systems for real short-video platforms. Moreover, the evaluation protocol considers multiple complementary aspects of restoration quality, including distortion fidelity~\cite{lu2024styleam}, perceptual quality~\cite{lpips,guan2024qmamba,yu2024sfiqa}, and temporal consistency~\cite{uav}. We expect this challenge to provide a useful testbed for the community and to encourage the development of more effective and practical generative restoration methods for S-UGC videos.


This challenge is one of the challenges associated with the NTIRE 2026 Workshop~\footnote{\url{https://www.cvlai.net/ntire/2026/}} on:
deepfake detection~\cite{ntire26deepfake}, 
high-resolution depth~\cite{ntire26hrdepth},
multi-exposure image fusion~\cite{ntire26raim_fusion}, 
AI flash portrait~\cite{ntire26raim_portrait}, 
professional image quality assessment~\cite{ntire26raim_piqa},
light field super-resolution~\cite{ntire26lightsr},
3D content super-resolution~\cite{ntire263dsr},
bitstream-corrupted video restoration~\cite{ntire26videores},
X-AIGC quality assessment~\cite{ntire26XAIGCqa},
shadow removal~\cite{ntire26shadow},
ambient lighting normalization~\cite{ntire26lightnorm},
controllable Bokeh rendering~\cite{ntire26bokeh},
rip current detection and segmentation~\cite{ntire26ripdetseg},
low light image enhancement~\cite{ntire26llie},
high FPS video frame interpolation~\cite{ntire26highfps},
Night-time dehazing~\cite{ntire26nthaze,ntire26nthaze_rep},
learned ISP with unpaired data~\cite{ntire26isp},
short-form UGC video restoration,
raindrop removal for dual-focused images~\cite{ntire26dual_focus},
image super-resolution (x4)~\cite{ntire26srx4},
photography retouching transfer~\cite{ntire26retouching},
mobile real-word super-resolution~\cite{ntire26rwsr},
remote sensing infrared super-resolution~\cite{ntire26rsirsr},
AI-Generated image detection~\cite{ntire26aigendet},
cross-domain few-shot object detection~\cite{ntire26cdfsod},
financial receipt restoration and reasoning~\cite{ntire26finrec},
real-world face restoration~\cite{ntire26faceres},
reflection removal~\cite{ntire26reflection},
anomaly detection of face enhancement~\cite{ntire26anomalydet},
video saliency prediction~\cite{ntire26videosal},
efficient super-resolution~\cite{ntire26effsr},
3d restoration and reconstruction in adverse conditions~\cite{ntire26realx3d},
image denoising~\cite{ntire26denoising},
blind computational aberration correction~\cite{ntire26aberration},
event-based image deblurring~\cite{ntire26eventblurr},
efficient burst HDR and restoration~\cite{ntire26bursthdr},
low-light enhancement: `twilight cowboy'~\cite{ntire26twilight},
and efficient low light image enhancement~\cite{ntire26effllie}.

\section{Challenge}
\label{sec:challenge}
The NTIRE 2026 Challenge on Short-form UGC Video Restoration in the Wild with Generative Models aims to advance practical restoration techniques for short-form UGC videos. In contrast to conventional restoration settings, S-UGC videos collected from real short-video platforms usually suffer from diverse and entangled degradations introduced by consumer-level acquisition, compression, transcoding, and platform-side processing pipelines. These characteristics make restoration in the wild particularly challenging, especially when both perceptual realism and temporal coherence need to be considered.

To support this challenge, we introduce the \textit{KwaiVIR} benchmark, which is contributed by USTC and the Kuaishou company. KwaiVIR contains both synthetically degraded videos and real-world S-UGC videos collected in the wild, so as to better reflect practical restoration scenarios on short-video platforms. The released training set contains 200 videos from the synthetic dataset and 48 videos from the wild dataset. In addition, the validation set contains 11 videos from the synthetic and wild datasets, while the test set contains 20 videos from the synthetic and wild datasets. By combining paired synthetic data and authentic real-world videos within a unified benchmark, this challenge is designed to encourage the development of video restoration methods that are not only faithful to reference content but also perceptually convincing in realistic applications.

Different from conventional restoration benchmarks that focus mainly on distortion fidelity, this challenge emphasizes a more comprehensive evaluation protocol for generative video restoration. In particular, the challenge considers both restoration accuracy and perceptual quality on synthetic data and perceptual quality on real-world data. Moreover, in this edition, separate rewards are provided for the objective track and the subjective track, so that participating methods can be evaluated under different optimization goals. This setting is particularly suitable for S-UGC video restoration, where fidelity-oriented restoration and perceptually preferred restoration may lead to different design choices.

The evaluation protocol of this challenge can be divided into two parts. The first part is an objective comparison, where we utilize PSNR, SSIM, LPIPS, MUSIQ and WarpError to obtain a final score for synthetic videos, while utilizing MUSIQ and WarpError for the wild videos. These two parts forms the final objective score. In the test stage, we only evaluated 11 videos in total due to the limited runtime in the Codabench platform~\cite{xu2022codabench}. For subjective comparison, we organize two professional students and workers from education and industry to conduct subjective experiments for all teams. The evaluation is achieved from three dimensions, including content fidelity, perceptual quality, and temporal consistency. These three dimensions result in the final subjective score. The videos for subjective comparison include 10 synthetic videos and 10 wild videos.

\section{Challenge Results}
\label{challenge_results}
The challenge results are summarized in Table~\ref{tab:results-subjective} and Table~\ref{tab:results-objective}, which present the subjective user study and objective evaluation results, respectively. Overall, RedMediaTech achieved the best performance, ranking first in both the subjective and objective evaluations. The subjective comparison is based on three dimensions, including fidelity, perception and temporal consistency. Team RedMediaTech obtained the highest subjective score of 3.8525. Team TaoMC2, STCVSR, and MiAlgo LM followed closely, all achieving subjective scores above 3.78, which indicates strong perceptual quality and favorable human preference. IMAG@NJUST and BuptMM also delivered competitive subjective performance, ranking fifth and sixth, respectively. 

For the objective evaluation, team RedMediaTech again ranked first with the highest final score of 61.7395, achieving the best PSNR (30.7610), SSIM (0.8504), and LPIPS (0.1910), which demonstrates a strong capability in preserving both fidelity and perceptual quality. Team Video-Restorer ranked second with an objective score of 61.5871 and achieved the lowest WarpError (0.0549), suggesting superior temporal consistency among all submitted videos. Team Lucky one, BuptMM, IMAG@NJUST, and TaoMC2 also achieved competitive objective performance, all obtaining final scores above 56.0.

It is worth noting that the subjective and objective rankings are not strictly aligned for all teams. This phenomenon is expected in the present challenge setting, as participants were allowed to submit different restoration results for the subjective and objective tracks. Consequently, the two rankings should be interpreted as reflecting performance under different optimization goals rather than as being directly comparable on a one-to-one basis. Nevertheless, the results still provide valuable evidence that different methods exhibit distinct advantages in perceptual quality, distortion fidelity, and temporal consistency, highlighting the importance of comprehensive evaluation protocols for generative video restoration.

\begin{table*}[]
\centering
\resizebox{0.6\textwidth}{!}{\begin{tabular}{c|c|c|c}
\hline
 Team name          & Team leader            & Subjective Score & \makecell{Ranking\\(User Study)}   \\ \hline
RedMediaTech & Yuxiang Chen &  3.8525 & 1 \\
TaoMC2 &  Jiahui Liu & 3.79875 & 2 \\ 
STCVSR & Junyang Chen &3.785 & 3 \\
MiAlgo LM &  Jiyuan Zhang & 3.781875 & 4 \\ 
IMAG@NJUST & Yixin Yang & 3.644375 & 5 \\
BuptMM & Xiuhao Qiu & 3.606875 & 6 \\
Lucky one & Jingyi Xu &3.5575 & 7 \\ 
Video-Restorer   &    Zhirui Liu &  3.49875  & 8    \\
BVI      & Yuxuan Jiang &  3.136875 & 9  \\
xingyanan    & Yanan Xing &  3.1325 & 10   \\
NTR     & Jiachen Tu  &  3.1175   & 11   \\
weichow & Wei Zhou  &  2.728125 & 12  \\
  \hline
\end{tabular}}
\caption{The rank of different teams in terms of subjective quality (Primary Track).}
\label{tab:results-subjective}
\end{table*}

\begin{table*}[]
\centering
\resizebox{0.95\textwidth}{!}{\begin{tabular}{c|c|c|ccccc|c}
\hline
 Team name          & Team leader            & \makecell{Score\\ (Objective)} & PSNR  & SSIM & LPIPS & MUSIQ &   WarpError &    \makecell{Ranking \\(Objective)}   \\ \hline
 
RedMediaTech & Yuxiang Chen &  61.7395 &	30.7610 &	0.8504 &	0.1910 &	67.7429 & 0.0628  &  1 \\
Video-Restorer          &    Zhirui Liu           & 61.5871 &	29.3481&	0.8273 &	0.2118 &	68.9456 &	0.0549 &  2 \\
Lucky one & Jingyi Xu & 57.6196 & 	29.3394 & 	0.8264 & 	0.2172 &62.8006 & 	0.0568 &  3 \\ 
BuptMM & Xiuhao Qiu & 56.6877 & 	29.3096 & 	0.8302 & 	0.1934 & 63.5165 & 	0.0654 &  4 \\ 
IMAG@NJUST & Yixin Yang & 56.3061 & 	28.1188 & 	0.8031 & 	0.2110 & 67.9077 & 	0.0644 & 5 \\ 
TaoMC2 &  Jiahui Liu &  56.2668 & 	27.7289 & 	0.7899 & 	0.2082 & 68.4946 & 	0.0633 & 6 \\ 
MiAlgo LM &  Jiyuan Zhang & 54.0163 & 	29.6824 & 	0.8203 & 	0.2406 & 62.0506 & 	0.0642 &  7 \\ 
       STCVSR & Junyang Chen &  53.3945 &	27.8981 &	0.7771 &	0.2309&	67.5804 &	0.0669  &  8 \\
xingyanan            & Yanan Xing                      & 49.7414 &	25.6962 &	0.7571 &	0.2429 &	70.1317 &	0.072 & 9 \\
 NTR                & Jiachen Tu              & 49.2197 &	28.7295 &	0.8121&	0.3161&	58.211&	0.0572  & 10 \\
weichow & Wei Zhou                       & 47.9321  & 	28.4218  & 	0.7932 & 	0.2793  & 	60.4755  & 	0.0685   & 11\\
BVI      & Yuxuan Jiang              & 46.7834 &	24.8536 &	0.7255 &0.3184 &	74.8298 &	0.0711 &  12 \\
  \hline
\end{tabular}}
\caption{The rank of different teams in terms of objective quality.}
\label{tab:results-objective}
\end{table*}

\section{Teams and Methods}
\label{sec:teams_and_methods_1}

\subsection{RedMediaTech}

\begin{figure}[t]
\centering
\includegraphics[width=1.0\linewidth]{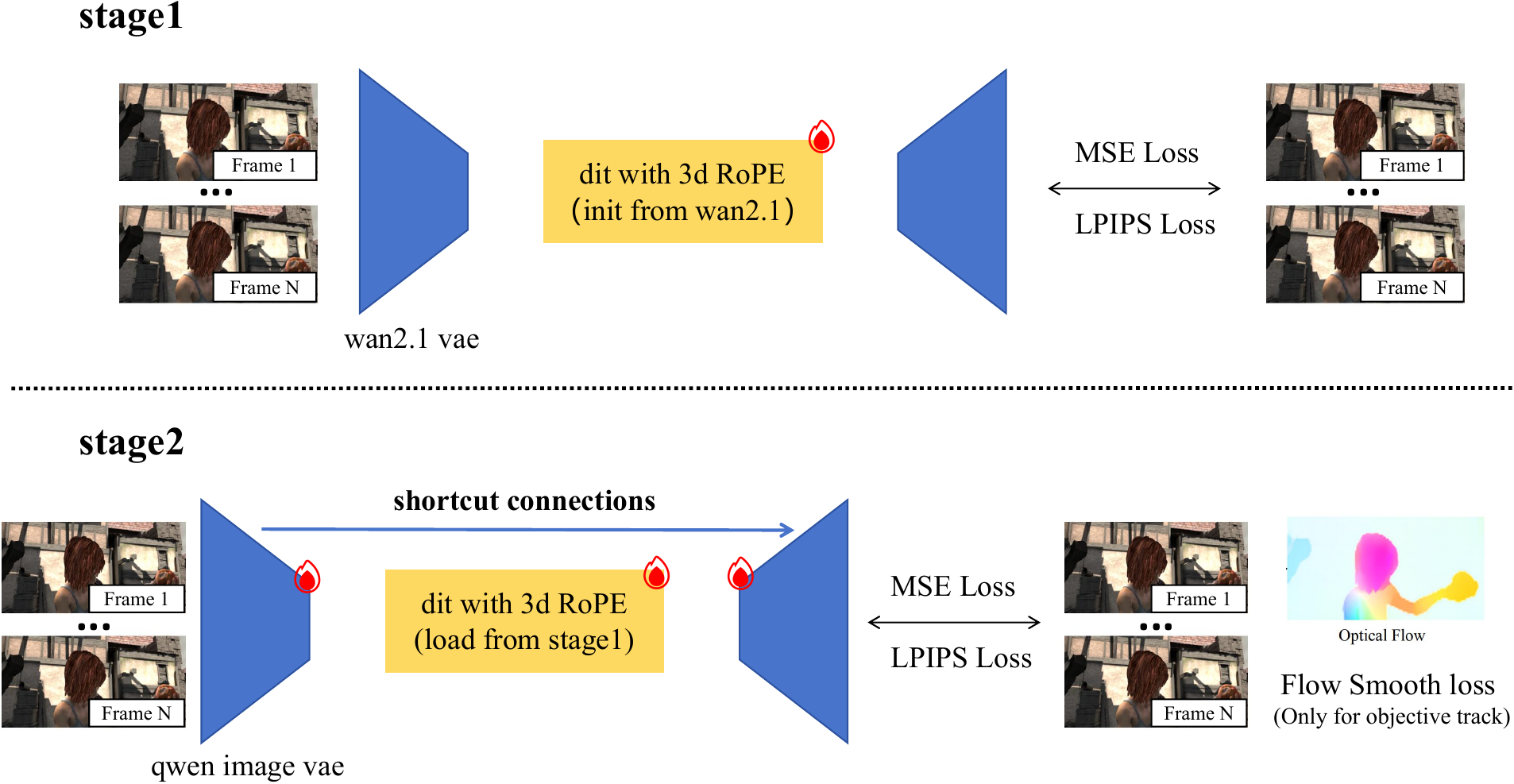}
\caption{The overall framework of Team RedMediaTech. }
\label{fig:RedMediaTech}
\end{figure}

Team RedMediaTech tackles the UGC video restoration challenge with a single‑step diffusion framework built on Wan 2.1’s diffusion transformer DiT~\cite{wan2025wan}. Their approach uses a two‑stage training scheme to balance perceptual quality and distortion‑oriented metrics. In the first stage, the model is initialized with the Wan 2.1 VAE and DiT and trained with a mixture of mean‑squared error and LPIPS losses~\cite{lpips}. This stage exploits the strong generative prior of Wan 2.1 to achieve rapid convergence and good perceptual quality. The second stage replaces the Wan 2.1 VAE with the more expressive Qwen‑Image VAE~\cite{wu2025qwenimage} and continues training to improve PSNR and SSIM. Shortcut connections between the VAEs help preserve spatial details, and the DiT adopts a three‑dimensional rotary positional encoding, often called RoPE~\cite{lazos2005rope}, to capture temporal information. Data augmentations such as temporal frame skipping and random pair cropping enhance robustness. The team also reports using an internal dataset of about 10 000 high‑resolution video clips in addition to the official NTIRE training set.

\noindent\textbf{Training details} The method is implemented in PyTorch with the \texttt{diffsynth‑studio} library. Stage 1 runs for approximately five days on eight H20 GPUs with 140 GB of memory using AdamW with betas $(0.9, 0.99)$ and a learning rate of $5\times 10^{-5}$. Stage 2 trains for one more day with a reduced learning rate of $2\times 10^{-5}$. The internal dataset complements the official NTIRE data and broadens the training distribution. Augmentations include temporal frame skipping and random cropping, and MSE and LPIPS losses remain active across both stages; flow‑smoothness regularization is applied for the objective track.

\noindent\textbf{Testing details} For each video, frames are encoded into the latent space via the trained VAE—Wan 2.1 in stage 1 or Qwen‑Image in stage 2—and a single diffusion step using the DiT with RoPE is applied. The cleaned latent representation is then decoded to produce the restored frame. This single‑step procedure yields efficient inference while maintaining high perceptual quality. The combination of three‑dimensional positional encoding and shortcut connections helps maintain temporal coherence. The final system relies solely on the student model; teacher or LoRA branches are not used, and frames are padded or cropped as needed to satisfy backbone constraints.

\subsection{TaoMC2}
 Their framework~\cite{ntire26TaoMC2} adopts a two-stage generative restoration pipeline based on text-to-video (T2V) diffusion models. In Stage 1, they design a dual-branch restoration module, where one branch focuses on general real-world restoration and the other incorporates a pre-cleaning module to handle discontinuities and severe degradations. For the objective track, the framework further leverages the open-source DOVE model~\cite{dove} as a complementary branch. In Stage 2, an RRDB-based fusion network~\cite{wang2018esrgan} combines the original degraded input with the intermediate outputs from both generative branches. This input-anchored fusion strategy improves robustness by balancing degradation removal and detail preservation, while mitigating over-smoothing and hallucination artifacts.

\noindent \textbf{Training details} The model is trained on both the official NTIRE 2026 synthetic dataset and an additional large-scale curated dataset. The official training set contains 200 synthetic videos, while the extra data consists of 500K high-quality text-video pairs selected from an initial pool of 3M web videos collected from sources such as YouTube and Pexels. These videos are filtered using no-reference video quality assessment models~\cite{DOVER,zhang2023md}, and text descriptions are generated using Qwen2.5-VL~\cite{bai2025qwen25vltechnicalreport}. During training, low-quality inputs are synthesized on-the-fly through a degradation pipeline including downsampling, noise, compression, and resizing. Stage 1 fine-tunes a CogVideoX1.5-based T2V diffusion backbone~\cite{yang2024cogvideox} with dual branches, while Stage 2 trains the RRDB-based fusion network. The training uses 49-frame clips cropped to $1024\times1024$, the AdamW optimizer, an initial learning rate of $2\times10^{-5}$, cosine annealing with warmup, and a distributed setup with 64 NVIDIA H20 GPUs. Pre-computed text embeddings are further adopted to reduce memory overhead.

\noindent \textbf{Testing details} During inference, the model performs single-step denoising without adding extra noise. For high-resolution videos, a patch-based aggregation strategy with direct block stitching is applied to avoid boundary artifacts. The outputs of the dual-branch generative stage are then fused by the RRDB-based refinement network to produce the final restored result. The method is evaluated on offline and official validation/test sets using perceptual, temporal, and full-reference metrics, including MUSIQ, WE, PSNR, SSIM, and LPIPS. No explicit ensemble is adopted in the final submission, as robustness is primarily achieved through the dual-branch design and the fusion strategy.

\subsection{STCVSR}

This team proposes a restoration pipeline based on the pretrained STCDiT~\cite{stcdit} and ODTSR~\cite{odtsr} models for short-form UGC video restoration. Their method aims to exploit the strong generative capability of diffusion-based video restoration while preserving fine structures and temporal consistency. The key idea is to use high-quality restored anchor frames together with suitable VAE-based segmented encoding, so that the final restoration process can better recover local structures and maintain coherent motion across frames.

Their method mainly consists of two components. First, they employ ODTSR~\cite{odtsr} to enhance sparse anchor frames sampled from the input video. These enhanced frames provide faithful structural guidance for subsequent video restoration. Second, they adopt STCDiT~\cite{stcdit} to restore the full video, leveraging its ability to model cross-segment relationships among clip latents encoded by a motion-aware VAE. For videos without dense textures, one frame every 25 frames is enhanced and used as an anchor to improve restoration quality. For videos with dense textures, they directly apply STCDiT~\cite{stcdit} without additional enhanced anchors, since discrepancies among enhanced frames may introduce temporal inconsistency.

To further improve robustness, they also handle challenging cases where local structures are severely degraded or suddenly lost in certain frames. In such cases, instead of using a fixed segmented encoding strategy, they identify the affected frames and adjust the segment boundaries for more suitable VAE-based segmented encoding. In this way, their method better preserves structure continuity and alleviates failure cases caused by inappropriate clip partitioning.

\noindent \textbf{Training details} This team does not conduct additional training for the challenge. Instead, they directly use the pretrained ODTSR~\cite{odtsr} and STCDiT~\cite{stcdit} models for inference. The method therefore relies on the pretrained weights and training protocols of the original models. No extra competition-specific finetuning or additional data are used in their final submission.

\noindent \textbf{Testing details} During testing, they first apply ODTSR~\cite{odtsr} to enhance sparse frames at an interval of 25 frames, and selectively replace the corresponding degraded frames with enhanced ones according to the video content. These enhanced frames are then used as additional anchors for STCDiT~\cite{stcdit}. After that, STCDiT~\cite{stcdit} is used to restore the whole video. For videos with dense textures, they skip the anchor enhancement strategy and directly use STCDiT~\cite{stcdit} in order to avoid temporal inconsistency. For particularly challenging videos with severely degraded local structures, they further adjust the segment boundaries in the VAE encoding stage to obtain better restoration results.

\subsection{MiAlgo\_LM}

\begin{figure}[t]
  \centering
   \includegraphics[width=\linewidth]{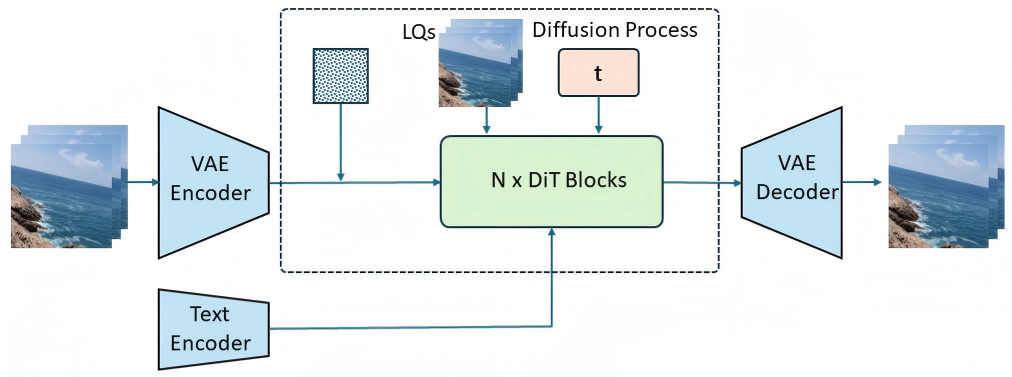}
   \caption{The proposed video restoration framework built upon the Wan2.1-14B Diffusion Transformer (DiT), named as Mi-GenVR.}
   \label{fig:head}
\end{figure}

\begin{figure}[t]
  \centering
   \includegraphics[width=\linewidth]{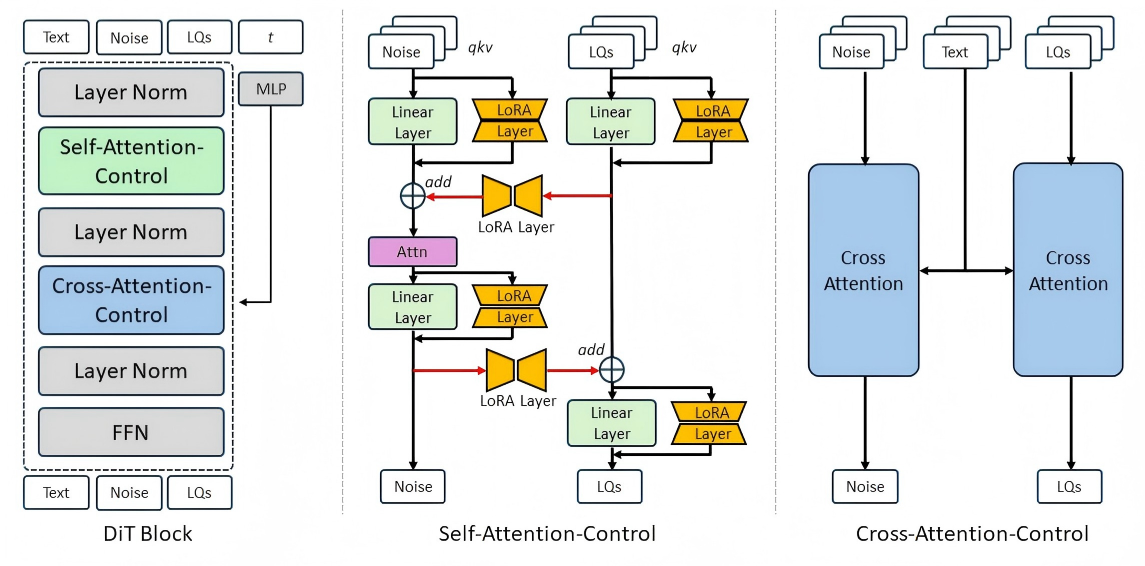}
   \caption{The DiT design of the proposed video restoration framework.}
   \label{fig:head2}
\end{figure}

This team proposes a method titled ``Video as Mini-Control for Generative Video Restoration Model''. Their framework, named Mi-GenVR, is built upon the Wan2.1-14B Diffusion Transformer~\cite{wan2025wan} and introduces a dual-stream conditional injection design for generative video restoration, as shown in Fig.~\ref{fig:head}. To bridge the gap between pretrained video generation models and high-fidelity restoration, the method explicitly incorporates low-quality (LQ) priors into the diffusion process through two proposed modules: Self-Attention-Control and Cross-Attention-Control, as shown in Fig~\ref{fig:head2}. In the Self-Attention-Control module, noisy latent features and degraded inputs are processed in parallel, and the LQ stream injects structural priors into the generative stream through bidirectional feature interaction. In the Cross-Attention-Control module, the text condition is decoupled into two parallel cross-attention paths, ensuring semantic consistency for both the generative stream and the LQ condition stream. To enable efficient adaptation of the 14B-parameter backbone, most pretrained weights are frozen and task-specific tuning is achieved mainly through high-rank LoRA modules.

\noindent \textbf{Training details} The framework is implemented in PyTorch and based on the pretrained Wan2.1-14B model~\cite{wan2025wan}. All LoRA modules use rank $r=512$ to balance parameter efficiency and adaptation capacity. During training, the model takes 9-frame clips as input and directly processes native high-definition videos at $1920\times1080$ resolution without downsampling. Training is conducted on 8 NVIDIA H200 GPUs with a total batch size of 8, using the AdamW optimizer and a learning rate of $1\times10^{-4}$. The model is trained for 9{,}000 steps. No extra data is used beyond the competition data.

\noindent \textbf{Testing details} During inference, the method adopts a temporal sliding-window strategy to process videos of arbitrary length. The temporal window size is set to 9 frames with a stride of 5 frames, and predictions of overlapping frames from adjacent windows are averaged to ensure seamless and temporally consistent restoration. For the final segment of each video, the window is shifted backward to maintain a complete 9-frame input. The inference process uses 50 denoising steps. The full model contains about 20.161B parameters, including the frozen 14.288B Wan2.1 backbone and 5.87B LoRA parameters, while maintaining a peak inference memory of 52.90 GB. Inference on a 9-frame $1088\times1920$ clip takes 224.28 seconds in total, corresponding to about 4.49 seconds per denoising step.

\subsection{Gen-VSR}

This team proposes Gen-VSR, which utilizes a sequential two-stage pipeline. Short-form UGC in the wild motivates this two-stage design: the method first reduces compression and noise to stabilize frames before applying generative super-resolution, matching typical mobile and capture artifacts. In the first stage, the pipeline loads degraded short-form UGC videos and restores each clip using a temporal model. This model performs alignment via optical flow or deformable alignment, incorporates optional pre-cleaning, and handles feature extraction. Subsequently, it reconstructs each frame with an ONNX runtime upsampler that is fed with low-resolution RGB inputs and the extracted features. To maintain temporal consistency, sliding temporal windows with a length of 30 and an overlap of 10 are utilized, and intermediate outputs are written as MP4 files. In the second stage, the method takes the Stage 1 outputs as input and runs DOVE~\cite{dove}, an official one-step diffusion Video Super-Resolution (VSR) model built on CogVideoX~\cite{yang2024cogvideox}. This approach combines explicit temporal restoration and upsampling with a state-of-the-art one-step real-world VSR diffusion model. It specifically targets sharpness and temporal coherence on noisy UGC without the latency typically associated with multi-step diffusion competitors. Final restored videos are generated as the Stage 2 outputs.

\noindent \textbf{Testing details} The team does not perform any training or finetuning on additional datasets for this submission, relying entirely on fixed pretrained checkpoints shipped with their code. The single deterministic pipeline avoids averaging artifacts and keeps the runtime predictable. The end-to-end complexity is heavily dominated by DOVE's large transformer and video diffusion backbone. The inference pipeline is implemented in Python, utilizing frameworks such as PyTorch, ONNX Runtime, Diffusers, Transformers, Decord/OpenCV, and FFmpeg. Stage 1 is tested with PyTorch 2.6+ with CUDA, while DOVE~\cite{dove} requires PyTorch $\ge 2.5$. Peak GPU memory usage is high for long or high-resolution clips; an NVIDIA GPU with $\ge 24$ GB is recommended for Stage 1. For the DOVE model, optional temporal chunking, spatial tiling, and CPU offload are employed to manage memory and prevent Out-Of-Memory errors.

\subsection{Lucky one}

Team Lucky one proposes an efficient single‑step diffusion‑based video restoration method built by fine‑tuning a pretrained CogVideoX model~\cite{yang2024cogvideox} using a latent‑pixel training strategy. Operating in the latent space, the model predicts the noise for each degraded frame in a single step, eliminating the need for multi‑step diffusion while still delivering high‑quality reconstruction. Latent‑pixel supervision leverages pixel‑level information in the latent domain to improve spatial fidelity and maintains temporal consistency across frames. According to their factsheet, this design achieves up to a 28× speed‑up compared with multi‑step diffusion models while preserving visual fidelity and temporal coherence. The authors note that no additional data beyond the official NTIRE 2026 training and validation sets is used.

\noindent\textbf{Training details} They fine‑tune the CogVideoX model for single‑step diffusion restoration using latent‑pixel supervision. Training pairs of degraded and clean frames from the official dataset teach the model to predict latent noise in one step. Although specific hyperparameters such as the optimizer, learning rate and GPU resources are not provided, the team stresses that the latent‑pixel design enables efficient optimization. They also experimented with previously published methods like DOVE and DGAFVSR as baselines~\cite{dove,xu2025rethinking}.

\noindent\textbf{Testing details} In inference, each degraded frame is encoded into latent space, a single diffusion step predicts the noise, and the latent representation is decoded back to the image domain. This frame‑wise procedure ensures fast inference and consistent restored videos. There is no multi‑step sampling, additional fusion or spatial upscaling; the method operates at the video’s original resolution.

\subsection{BuptMM}

Team BuptMM constructs a two‑branch restoration pipeline by combining the large diffusion transformer model SeedVR2 7B~\cite{wang2025seedvr2} with the lighter video super‑resolution model FlashVSR~\cite{zhuang2025flashvsr}. Both branches rely on diffusion transformer architectures. The SeedVR2 branch is fine‑tuned for the challenge using Low‑Rank Adaptation (LoRA)~\cite{hu2022lora} to efficiently adapt the pretrained model to the NTIRE data, while the FlashVSR branch emphasizes perceptual quality and temporal stability. Each input video is processed independently by the two branches, and their outputs are subsequently aligned in spatial resolution and temporal length. The team then performs a \emph{full‑frame unified fusion} by applying a constant linear weight across all pixels, avoiding region‑wise adjustments. Through ablation experiments they select a fusion weight $\alpha=0.7$; larger values favour SeedVR2 and improve distortion metrics such as PSNR, whereas smaller values favour FlashVSR and enhance perceptual scores.

\noindent\textbf{Training details} The SeedVR2 7B branch is fine‑tuned on the official NTIRE 2026 dataset without external data. The network is optimized with AdamW at an initial learning rate of $8\times 10^{-5}$ for 30 epochs on eight NVIDIA H200 GPUs, and spatial tiling is used to match the optimal resolution range of SeedVR2 and reduce memory consumption. FlashVSR retains its default weights. No additional data augmentation or synthetic degradation is reported.

\noindent\textbf{Testing details} During inference, video frames are processed by both branches in parallel. The aligned outputs are fused using the constant weight $\alpha=0.7$, avoiding block‑wise or channel‑wise weighting and thus preventing boundary artefacts and temporal jitter. Spatial tiling and batch inference are employed to reduce the memory footprint and improve throughput. The resulting method combines the high pixel‑level fidelity of SeedVR2 with the perceptual quality and temporal consistency offered by FlashVSR.

\subsection{IMAG@NJUST}

This team proposes a diffusion-based video restoration framework centered on CoDiVSR, titled ``Rethinking What to Condition and What to Disentangle in Diffusion-based Video Super-Resolution''. For the objective track, they adopt a single-stage diffusion-based approach built upon the CogVideoX1.5-5B-I2V architecture. The method focuses on improving video super-resolution by redesigning conditioning and disentanglement strategies within the diffusion process. For the subjective track, they employ a two-stage pipeline that first enhances video quality using Vivid-VR~\cite{bai2026vividvr}, followed by refinement using the CoDiVSR model.

\noindent \textbf{Training details} The method leverages pretrained large-scale generative models, including CogVideoX1.5-5B-I2V and PLLaVa-13B for caption extraction. Additional data is incorporated through automatically generated captions using PLLaVa-13B, which provide textual conditioning for the diffusion model. Detailed training strategies are not fully specified in the factsheet, but the approach follows the standard CogVideoX-based diffusion training pipeline, as described in the official CoDiVSR implementation.

\noindent \textbf{Testing details} During inference, the objective-track model directly applies the pretrained CoDiVSR model using the CogVideoX pipeline with a DPMScheduler. The inference is performed with 50 sampling steps on a single NVIDIA H20 GPU, with memory optimizations such as VAE slicing, tiling, and sequential CPU offloading. For the subjective track, the pipeline first processes inputs using Vivid-VR and then refines them with CoDiVSR. This two-stage strategy improves perceptual quality compared to single-stage inference.

\subsection{BVI}
\begin{figure}[t]
  \centering
  \includegraphics[width=1.0\linewidth]{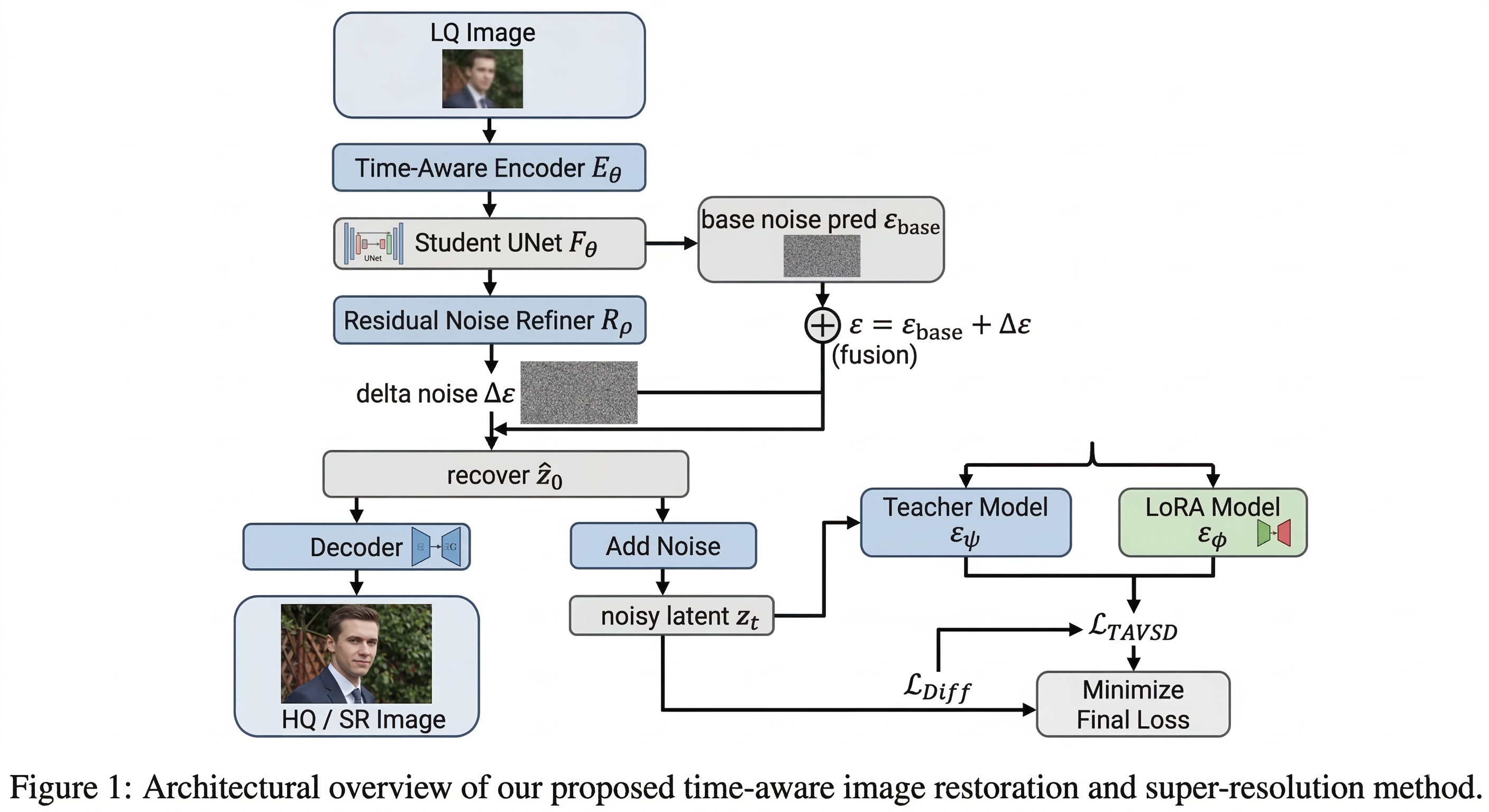} 
  \caption{The overall framework of Team BVI. }
  \label{fig:BVI}
\end{figure}
This team builds its solution upon the \textbf{Time-Aware one-step Diffusion Network for real-world image super-resolution (TADSR)} framework~\cite{tadsr} and adapts it to short-form UGC video restoration. Their method follows a one-step latent diffusion paradigm and processes videos in a frame-wise manner. For each degraded frame, the input is first encoded into a timestep-aware latent representation, then a student branch predicts the latent noise in a single step, and finally the restored result is decoded back to image space. This design preserves the efficiency advantage of one-step diffusion while maintaining strong restoration capability for real-world video content.

On top of the original TADSR framework~\cite{tadsr}, they introduce several modifications to improve restoration quality. First, they add a lightweight residual noise refiner to the one-step student branch, which helps refine the latent prediction and improve restoration details. Second, they propose a detail-aware training strategy that explicitly emphasizes high-frequency and gradient information. In addition to the original reconstruction and perceptual supervision, they introduce weighted Charbonnier losses on both high-frequency residuals and image gradients. A detail-weight map is constructed from reconstruction error, target-image gradients, and high-frequency residuals, and these detail-aware losses are progressively activated with a warm-up schedule so that the model first learns stable global restoration and then gradually focuses on local refinement.

They also introduce a ratio-capped residual regularization mechanism for stable training. Instead of directly penalizing the residual correction magnitude, they constrain the ratio between the residual correction and the base prediction, and only activate this regularization when the ratio exceeds a scheduled target. This prevents the residual refiner from being overly suppressed while still keeping it stable. Overall, their method is a modified one-step latent diffusion restoration framework that combines efficiency, perceptual quality, and detail enhancement.

\noindent \textbf{Training details} The training strategy follows the one-step TADSR paradigm~\cite{tadsr}. For each degraded-clean pair, the degraded frame is encoded into a timestep-aware latent space, and the student branch predicts the latent noise in one step. The restored output is supervised by the original TADSR objectives together with LPIPS loss~\cite{lpips}, detail-aware high-frequency loss, gradient detail loss, and ratio-capped residual regularization. The LoRA branch is optimized with a diffusion loss during training~\cite{lora}. In addition to the official NTIRE 2026 training and validation data, they use extra data from LSDIR~\cite{li2023lsdir} and BVI-AOM~\cite{nawala2024bvi}, and optionally synthetic degraded-clean pairs generated with Real-ESRGAN-style degradation~\cite{realesrgan}. Their implementation is based on PyTorch and diffusers, and they use AdamW with an initial learning rate of $5\times10^{-5}$. The report indicates training on 4 GH200 GPUs for about 10 hours with 100 epochs in total.

\noindent \textbf{Testing details} During testing, the input video is processed frame by frame. For each degraded frame, they perform timestep-aware latent encoding, one-step student noise prediction, residual noise refinement, latent reconstruction, and VAE decoding. The teacher model and LoRA branch are discarded during inference. To satisfy the spatial constraints of the latent diffusion backbone, the input frame is padded or resized when necessary, and the restored output is finally cropped or resized back to the original resolution. Their final system performs restoration without spatial upscaling, uses one sampling step, and reports an inference time of about 0.169 s per image.

\subsection{xingyanan}

This team proposes \textbf{FlashVSR-UGC-Causal}, a real-time one-step video restoration framework for in-the-wild short-form UGC videos. Their method is built upon the FlashVSR framework and is designed to address common degradations in UGC videos, including severe compression artifacts, low-light noise, and motion blur. By adopting distribution matching distillation (DMD), they compress the original multi-step diffusion process into a one-step restoration mapping, which substantially improves inference efficiency while preserving strong perceptual restoration capability.

To better adapt the method to practical UGC scenarios, they further introduce several system-level improvements. First, they design a vertical-stream and resolution adaptation strategy for portrait-format videos, where the output is aligned to a standard high-resolution format and tiled processing is adopted for memory-efficient inference on large frames. Second, they employ a causal projection strategy, where the restoration process depends only on the current frame and historical information rather than future frames. This causal design improves robustness under nonlinear motion and shot changes, and also helps avoid temporal artifacts caused by future-frame leakage. In addition, exponential moving average states are propagated across DiT blocks to enhance temporal consistency and reduce flickering artifacts.

They also introduce degradation-aware preprocessing to improve restoration stability. Specifically, the pipeline performs artifact suppression for common H.264/H.265 compression distortions before restoration, and an explicit color correction module is applied to align the restored output with the color distribution of the low-quality input. Combined with mixed-precision inference and dynamic memory management, their method achieves an efficient and practical restoration pipeline for large-scale UGC video processing.

\noindent \textbf{Training details} Their method is based on the FlashVSR framework and its DMD-based one-step diffusion design. The technical report mainly focuses on the inference system and practical deployment strategy rather than providing detailed challenge-specific retraining settings. The restoration framework relies on the pretrained FlashVSR implementation and integrates resolution adaptation, causal streaming, explicit color correction, and memory optimization for the NTIRE 2026 challenge setting.

\noindent \textbf{Testing details} During testing, the input video is first analyzed for its spatial resolution and aspect ratio. For portrait videos, the output is aligned to the target resolution, and tiled inference with overlap and Gaussian fusion is used when necessary to reduce memory consumption. Each frame is then restored with the one-step FlashVSR pipeline under a causal streaming setting, where only current and historical information are used. Artifact suppression and explicit color calibration are further applied to improve visual fidelity. Their optimized inference pipeline uses mixed precision and memory release strategies, and the report states that the method achieves about 17.2 FPS at 720p and 8.5 FPS at 1080p tiled inference on modern GPUs.

\subsection{NTR}

This team proposes a method titled ``Frame-wise UGC Video Restoration via Perceptual Fine-tuning of a Diffusion-Pretrained U-Net''. They adopt a two-stage training paradigm. In Stage 1, they pretrain a time-conditioned U-Net (TimeDiffiT) containing 142.4M parameters via Masked Diffusion Autoencoding (MDAE)~\cite{he2022masked} on a large-scale image corpus. The architecture utilizes a TimeDiffiT-U-Net encoder-decoder with four spatial scales, incorporating time conditioning via sinusoidal encoding, a 2-layer MLP AdaGN, and TimeAttention blocks across all scales. The MDAE~\cite{he2022masked} pretraining applies dual corruption: spatial masking of $p_{mask}\sim\mathcal{U}(1\%,75\%)$ on non-overlapping $16\times16$ blocks, and VE-SDE~\cite{song2020score} noise injection. This self-supervised pretraining provides a robust initialization without the need for paired labels. In Stage 2, the full encoder-decoder is fine-tuned on the competition's 200 paired synthetic videos, which consist of 32,400 frames. This fine-tuning uses a multi-loss objective that incorporates perceptual losses (VGG~\cite{simonyan2014very} and LPIPS) in a continuation phase after an initial MSE-only training phase, effectively reducing LPIPS without catastrophically degrading PSNR.

\noindent \textbf{Training details} The framework is implemented in PyTorch 2.0+ and optimized using AdamW. Stage 2a fine-tunes the MDAE-pretrained model with an MSE loss for 300 epochs. This stage utilizes $512\times512$ random crops, a batch size of 8 distributed across 2 NVIDIA H200 GPUs, and a learning rate of $5\times10^{-6}$. Time conditioning is achieved via an estimated noise level derived from scikit-image, and an Exponential Moving Average (EMA) is applied for checkpoint selection. Stage 2b (perceptual fine-tuning) resumes from the best MSE checkpoint. It continues training for 25 additional epochs using a combined loss of $\mathcal{L}_{MSE}+1.0\cdot\mathcal{L}_{VGG}+4.0\cdot\mathcal{L}_{LPIPS}$. In this stage, the learning rate is adjusted to $2\times10^{-6}$ and the batch size is reduced to 4 to manage the memory overhead introduced by the perceptual losses.

\noindent \textbf{Testing details} The learned model processes each video independently using frame-wise single-step inference, bypassing iterative diffusion sampling. It employs tiled processing with $512\times512$ patches, a stride of 384, and overlap blending. During inference, the per-tile noise level is estimated via scikit-image and fed as the time conditioning input. The method is applied identically to both synthetic and wild clips using a single GPU. The team does not use any ensembles or test-time augmentation in their final submission, noting that geometric self-ensembling degraded perceptual details. Finally, the restored output videos are encoded with the mp4v codec at their original frame rate.

\subsection{weichow}

This team proposes NAFNet-GAN, a lightweight video restoration method utilizing adversarial training. The approach employs a NAFNet-style~\cite{chen2022simple} U-Net architecture containing 12.5M parameters, configured with a width of 32, encoder blocks of [2, 2, 4, 8], and decoder blocks of [2, 2, 2, 2]. The model operates on the principle of residual learning, where the final output is obtained by adding the predicted residual to the input. This single-stage, per-frame restoration approach aims to balance fidelity metrics with perceptual metrics, achieving a good PSNR of 28.42 and a reasonable LPIPS of 0.279.

\noindent \textbf{Training details} The model is trained from scratch exclusively on the KwaiVIR synthetic training data, which consists of 200 video pairs and 36,000 frame pairs. Training frames are pre-extracted as PNG files for fast I/O. The implementation uses PyTorch 2.5.1 and optimizes the network using AdamW with a weight decay of 1e-4. The training is conducted on two NVIDIA H800 80GB GPUs using DataParallel. The training process is divided into two distinct stages. In the first stage, base training is performed for 5 epochs with a batch size of 8, a patch size of 256, and a learning rate of 2e-4 with cosine annealing. The loss function in this stage is a weighted sum of L1, 0.1 $\times$ VGG, and 2.0 $\times$ LPIPS losses. The second stage involves adversarial fine-tuning for 2 epochs by incorporating an additional UNet PatchGAN discriminator with 1.77M parameters. This stage utilizes spectral normalization~\cite{miyato2018spectral} and a hinge GAN loss to further enhance perceptual quality, setting the generator learning rate to 1e-4 and the discriminator learning rate to 4e-4. The total training time is approximately 2 hours.

\noindent \textbf{Testing details} At inference, the method performs per-frame processing at the native 1080x1920 resolution using 512px tile processing. No ensembles are used in the final submission. The final output videos are encoded using libx264 with a Constant Rate Factor (CRF) of 10. The inference process takes approximately 40 seconds per 92-frame video on an NVIDIA H800 80GB GPU.

\section*{Acknowledgments}
This challenge is sponsored by Kuaishou Technology. This work was partially supported by Grants of NSFC U25B2010, 62371434 and 623B2098, the Postdoctoral Fellowship Program of CPSF under Grant Number GZC20252293, China Postdoctoral Science Foundation-Anhui Joint Support Program under Grant Number 2024T017AH, China Postdoctoral Science Foundation under Grant Number  2025M783529, Anhui Postdoctoral Scientific Research Program Foundation (No.2025A1015), the Fundamental Research Funds for the Central Universities (No. WK2100250064).  This work was also partially supported by the Humboldt Foundation. We thank the NTIRE 2026 sponsors: OPPO, Kuaishou, and the University of Wurzburg (Computer Vision Lab).

\appendix
\section{Teams and Affiliations}
\subsection*{NTIRE2026 Organizers}

\noindent  \textit{\textbf{Title:}} NTIRE 2026 Challenge on Short-form UGC Video Restoration in the Wild with Generative Models: Datasets, Methods and Results

\noindent  \textit{\textbf{Members:}}

\noindent Xin Li\textsuperscript{1} (\textcolor{magenta}{xin.li@ustc.edu.cn})

\noindent Jiachao Gong\textsuperscript{2} (\textcolor{magenta}{gongjiachao@kuaishou.com})

\noindent Xijun Wang\textsuperscript{1} (\textcolor{magenta}{xijunwang00@gmail.com}) 

\noindent Shiyao Xiong\textsuperscript{2} (\textcolor{magenta}{xiongshiyao@kuaishou.com})

\noindent Bingchen Li\textsuperscript{1} (\textcolor{magenta}{lbc31415926@mail.ustc.edu.cn})

\noindent Suhang Yao\textsuperscript{1} (\textcolor{magenta}{yshustc@mail.ustc.edu.cn})

\noindent Chao Zhou\textsuperscript{2} (\textcolor{magenta}{zhouchao@kuaishou.com})

\noindent Zhibo Chen\textsuperscript{1} (\textcolor{magenta}{chenzhibo@ustc.edu.cn})

\noindent Radu Timofte\textsuperscript{3} (\textcolor{magenta}{radu.timofte@uni-wuerzburg.de})

\noindent  \textit{\textbf{Affiliations:}}

\noindent  \textsuperscript{1} University of Science and Technology of China

\noindent  \textsuperscript{2} KuaiShou Technology 

\noindent  \textsuperscript{3} Computer Vision Lab, University of Wurzburg, Germany

\subsection*{RedMediaTech}

\noindent  \textit{\textbf{Title:}} UGC Video Restoration using One Step Wan2.1~\cite{wan2025wan} 

\noindent  \textit{\textbf{Members:}}

\noindent   Yuxiang Chen (\textcolor{magenta}{chenyuxiang@xiaohongshu.com}), Shibo Yin, Yilian Zhong, Yushun Fang, Xilei Zhu, Yahui Wang, Chen Lu 

\noindent  \textit{\textbf{Affiliations:}}

\noindent Xiaohongshu Inc

\subsection*{TaoMC2}
\noindent Title: The Trade-off between Texture Preservation and Denoising: A Robust Dual-Branch Diffusion Pipeline for Challenging Short-Form Videos

\noindent  \textit{\textbf{Members:}} Meisong Zheng (\textcolor{magenta}{zhengmeisong.zms@alibaba-inc.com}, Xiaoxu Chen, Jing Yang, Zhaokun Hu, Jiahui Liu (\textcolor{magenta}{ljh263654@alibaba-inc.com}), Ying Chen, Haoran Bai, Sibin Deng, Shengxi Li, Mai Xu

\noindent  \textit{\textbf{Affiliations:}}

\noindent Alibaba Group Taobao \& Tmall Group, Beihang University

\subsection*{STCVSR}

\noindent  \textit{\textbf{Title:}} Accurate VAE Reconstruction and Faithful Structural Guidance Enable High-Quality Video Restoration 

\noindent  \textit{\textbf{Members:}}
\noindent   Junyang Chen (\textcolor{magenta}{jychen@njust.edu.cn}), Hao Chen, Xinzhe Zhu, Fengkai Zhang, Long Sun, Yixing Yang, Xindong Zhang, Jiangxin Dong, Jinshan Pan

\noindent  \textit{\textbf{Affiliations:}}

\noindent Nanjing University of Science and Technology, Hunan University, OPPO Research Institute

\subsection*{MiAlgo\_LM}

\noindent  \textit{\textbf{Title:}} Video as Mini-Control for Generative Video Restoration Model

\noindent  \textit{\textbf{Members:}}
\noindent   Jiyuan Zhang (\textcolor{magenta}{zhangjiyuan3@xiaomi.com}), Shuai Liu, Yibin Huang, Xiaotao Wang, Lei Lei

\noindent  \textit{\textbf{Affiliations:}}

\noindent Xiaomi

\subsection*{Video-Restorer}

\noindent  \textit{\textbf{Title:}} VideoSR

\noindent  \textit{\textbf{Members:}}
\noindent Zhirui Liu (\textcolor{magenta}{liuzhr28@mail2.sysu.edu.cn}), Shinan Chen, Shang-Quan Sun, Wenqi Ren

\noindent  \textit{\textbf{Affiliations:}}

\noindent Sun Yat-sen University, Xi'an Jiaotong University, Nanyang Technological University

\subsection*{Lucky one}

\noindent  \textit{\textbf{Title:}} Lucky one

\noindent  \textit{\textbf{Members:}}
\noindent   Jingyi Xu (\textcolor{magenta}{jingyixu@buaa.edu.cn}), Zihong Chen, Zhuoya Zou

\noindent  \textit{\textbf{Affiliations:}}

\noindent Beihang University, Tsinghua University

\subsection*{BuptMM}

\noindent  \textit{\textbf{Title:}} None

\noindent  \textit{\textbf{Members:}}
\noindent   Xiuhao Qiu (\textcolor{magenta}{18210287051@163.com}), Jingyu Ma, Huiyuan Fu, Kun Liu, Huadong Ma, Dehao Feng, Zhijie Ma, Boqi Zhang, Jiawei Shi, Hao Kang

\noindent  \textit{\textbf{Affiliations:}}

\noindent Beijing University of Posts and Telecommunications, JD Inc., China, China University of Petroleum

\subsection*{IMAG@NJUST}

\noindent  \textit{\textbf{Title:}} CoDiVSR: Rethinking What to Condition and What to Disentangle in Diffusion-based Video Super-Resolution

\noindent  \textit{\textbf{Members:}}
\noindent   Yixin Yang (\textcolor{magenta}{yangyixin@njust.edu.cn}), Yeying Jin, Junyang Chen, Hao Chen, Long Sun, Xu Cheng, Jinshan Pan

\noindent  \textit{\textbf{Affiliations:}}

\noindent Nanjing University of Science and Technology, National University of Singapore, Hunan University, Tsinghua University

\subsection*{BVI}

\noindent  \textit{\textbf{Title:}} Team BVI 

\noindent  \textit{\textbf{Members:}}
\noindent   Yuxuan Jiang (\textcolor{magenta}{dd22654@bristol.ac.uk}), Chengxi Zeng, Tianhao Peng, Fan Zhang, and David Bull 

\noindent  \textit{\textbf{Affiliations:}}

\noindent University of Bristol 

\subsection*{xingyanan}

\noindent  \textit{\textbf{Title:}} FlashVSR-UGC-Causal: Real-time One-step Video Restoration with Causal Streaming for In-the-wild Short-form UGC

\noindent  \textit{\textbf{Members:}}
\noindent   Yanan Xing (\textcolor{magenta}{18434795112@163.com})

\noindent  \textit{\textbf{Affiliations:}}

\noindent  Taiyuan University of Technology

\subsection*{NTR}

\noindent  \textit{\textbf{Title:}} Frame-wise UGC Video Restoration via Perceptual Fine-tuning of a Diffusion-Pretrained U-Net

\noindent  \textit{\textbf{Members:}}
\noindent   Jiachen Tu (\textcolor{magenta}{jtu9@illinois.edu}), Guoyi Xu, Yaoxin Jiang, Jiajia Liu, Yaokun Shi

\noindent  \textit{\textbf{Affiliations:}}

\noindent University of Illinois Urbana Champaign
\subsection*{weichow}

\noindent  \textit{\textbf{Title:}} NAFNet-GAN: Lightweight Video Restoration with Adversarial Training

\noindent  \textit{\textbf{Members:}}
\noindent   Wei Zhou (\textcolor{magenta}{weichow@u.nus.edu}), Linfeng Li, Hang Song, Qi Xu 

\noindent  \textit{\textbf{Affiliations:}}

\noindent National University of Singapore, Xi'an Jiaotong University, Shanghai Jiao Tong University















{
    \small
    \bibliographystyle{ieeenat_fullname}
    \bibliography{main}
}

\end{document}